\documentclass[a4paper]{article}

\usepackage{spconf}
\usepackage[utf8x]{inputenc}

\usepackage{stfloats}
\usepackage{mathtools}

\usepackage{array}
\usepackage{parskip}
\usepackage{graphicx}
\usepackage[table]{xcolor}
\usepackage{tikz}
\usepackage{hyperref}
\usepackage{soul}
\makeatletter
\def\thickhline{%
  \noalign{\ifnum0=`}\fi\hrule \@height \thickarrayrulewidth \futurelet
   \reserved@a\@xthickhline}
\def\@xthickhline{\ifx\reserved@a\thickhline
               \vskip\doublerulesep
               \vskip-\thickarrayrulewidth
             \fi
      \ifnum0=`{\fi}}
\makeatother

\newlength{\thickarrayrulewidth}
\usepackage{multirow}
\usepackage{adjustbox}

\title{End-to-end Speech Recognition: A review for the French language}
\name{Florian Boyer$^1$$^,$$^2$, Jean-Luc Rouas$^2$}
\address{
$^1$ Airudit, ENSEIRB-MATMECA, Talence, France\\
  $^2$ Univ. de Bordeaux, LaBRI, INP, CNRS, UMR5800, Talence, France\\
  \textit{florian.boyer@\{labri.fr, ea4t.com\}, jean-luc.rouas@labri.fr}\vspace{5mm}}
\begin{document}

\maketitle

\begin{abstract}
\vspace{0.3cm}
Recently, end-to-end ASR based either on sequence-to-sequence networks or on the CTC objective function gained a lot of interest from the community, achieving competitive results over traditional systems using robust but complex pipelines.
One of the main features of end-to-end systems, in addition to the ability to free themselves from extra linguistic resources such as dictionaries or language models, is the capacity to model acoustic units such as characters, subwords or directly words; opening up the capacity to directly translate speech with different representations or levels of knowledge depending on the target language.
In this paper we propose a review of the existing end-to-end ASR approaches for the French language. We compare results to conventional state-of-the-art ASR systems and discuss which units are more suited to model the French language.
\end{abstract}

\section{Introduction}

Automatic Speech Recognition (ASR) has traditionally used Hidden Markov Models (HMM), describing temporal variability, combined with Gaussian Mixture Models (GMM), computing emission probabilities from HMM states, to model and map acoustic features to phones.
In recent years, the introduction of deep neural networks replacing GMM for acoustic modeling showed huge improvements compared to previous state-of-the-art systems \cite{Graves13-SPW, Hinton14-DNN}. However, building and training such systems can be complex and a lot of preprocessing steps are involved. Traditional ASR systems are also factorized in several modules, the acoustic model representing only one of them along with lexicon and language models.

Recently, more direct approaches -- called end-to-end methods -- in which neural architectures are trained to directly model sequences of features as characters have been proposed \cite{Graves14-TET, Hannun14-DSS, Seide11-CST}. Predicting context independent targets such as characters using a single neural network architecture, drained a lot of interest from the research community as well as non-experts developers. This is caused by the simplicity of the pipeline and the possibility to create a complete ASR system without the need for expert knowledge.
Moreover having an orthographic-based output allows to freely construct words, making it interesting against the \textit{Out-Of-Vocabulary} problem encountered in traditional ASR systems.

End-to-end systems are nowadays extensively used and studied for multiple tasks and languages such as English, Mandarin or Japanese. However, for a language such as French, ASR performance and results with the existing methods have been scarcely studied, although the large number of silents letters, homophones or argot make comparing the assumptions made by each method very attractive.

In this context, we decided to study the three main types of architectures which have demonstrated promising results over traditional systems: 1) Connectionist Temporal Classification (CTC) \cite{Graves06-CTC, Miao15-EES} which uses Markov assumptions (\textit{i.e.} conditional independence between predictions at each time step) to efficiently solve sequential problems by dynamic programming, 2) Attention-based methods \cite{Chorowski15-ABM, Bahdanau16-ETE} which rely on an attention mechanism to perform non-monotonic alignment between acoustic frames and recognized acoustic units and 3) RNN-tranducer \cite{Graves13-SPW, Graves12-STW, Battenberg17-ENT} which extends CTC by additionally modeling the dependencies between outputs at different steps using a prediction network analogous to a language model.
We extend our experiments by adding two hybrid end-to-end methods: a multi-task method called joint CTC-attention \cite{Kim17-JCA, Hori17-AIJ} and a RNN-transducer extended with attention mechanisms \cite{Pra17-CSS}. To complete our review, we build a state-of-art phone-based system based on lattice-free MMI criterion \cite{Povey16-PST} and its \textit{end-to-end} counterpart with both phonetic and orthographic units \cite{Hadian18-ETE}.

\section{End-to-end systems for Speech Recognition}

	\subsection{Connectionist Temporal Classification}

The CTC \cite{Graves06-CTC} can be seen as a direct translation of conventional HMM-DNN ASR systems into \textit{lexicon-free} systems. Thus, the CTC follows the general ASR formulation, training the model to maximize $P(Y|X)$ the probability distribution over all possible label sequences:
    \begin{align*}
    	\hat{Y} = arg \max\limits_{Y\in\, \mathcal{A*}} p(Y|X)
    \end{align*}
    Here, $X$ denotes the observations, $Y$ is a sequence of acoustic units of length $L$ such that $Y = \{y_{l} \in\, \mathcal{A} | l = 1, ..., L\}$, where $\mathcal{A}$ is an \textit{alphabet} containing all distinct units.
As in traditional HMM-DNN systems, the CTC model makes conditional independence assumptions between output predictions at different time steps given aligned inputs and it uses the probabilistic chain rule to factorize the posterior distribution $p(Y|X)$ into three distributions (\textit{i.e.} framewise posterior distribution, transition probability and prior distribution of units).
However, unlike HMM-based models, the framewise posterior distribution is defined here as a framewise acoustic unit sequence $B$ with an additional blank label ${<}blank{>}$ such as $B = \{b_{t} \in\, \mathcal{A}\, \cup\, {<}blank{>} | t = 1, ..., T\}$.
    \begin{align*}
    	p(Y|X) = \underbrace{\sum\limits_{b=1}^{B}\prod\limits_{t=1}^{T} p(b_{t} | b_{t-1}, Y) p(b_{t}|X)}_{p_{\textrm{ctc}}(Y|X)} p(Y)
    \end{align*}
    
 Here, ${<}blank{>}$ introduces two contraction rules for the output labels, allowing to repeat or collapse successive acoustic units.

    \subsection{Attention-based model}\label{attention}
    
As opposed to CTC, the attention-based approach \cite{Chorowski15-ABM, Bahdanau16-ETE} does not assume conditional independence between predictions at different time steps and does not marginalize over all alignments.
Thus the posterior distribution $p(Y|X)$ is directly computed by picking a soft alignment between each output step and every input step as follows:
    \begin{align*}
        	p_{\textrm{att}}(Y|X) = \prod\limits_{l=1}^{U} p(y_{l} | y_{1}, ..., y_{l-1}, X)
    \end{align*} 
Here $p(y_{l}|y_{1},...,y_{l-1}, X)$, -- our attention-based objective function --, is obtained according to a probability distribution, typically a softmax, applied to the linear projection of the output of a recurrent neural network (or long-short term memory network), called decoder, such as:
    \begin{align*}
        p(y_{l}|y_{1},...,y_{l-1}, X) = \textrm{softmax}(\textrm{lin}(\textrm{RNN}(\cdot)))
    \end{align*}
The decoder output is conditioned by the previous output $y_{l-1}$, a hidden vector $d_{l-1}$ and a context vector $c_{l}$. Here $d_{l-1}$ denotes the high level representation (\textit{i.e.} hidden states) of the decoder at step $l-1$, encoding the target input, and $c_{l}$ designate the context -- or symbol-wise vector in our case -- for decoding step $l$, which is computed as the sum of the complete high representation $h$ of another recurrent neural network, encoding the source input $X$, weighted by $\alpha$ the attention weight:
		\begin{align*}
            c_{l} = \sum\limits_{s=1}^{S} \alpha_{l, s} h_{s} \text{ \quad,\quad }
            \alpha_{l, s} = \frac{\exp(e_{t, s})}{\sum\limits_{s'=1}^S \exp(e_{l, s'})}
        \end{align*}
where $e_{t}$, also referred to as \textit{energy}, measures how well the inputs around position $s$ and the output at position $l$ match, given the decoder states at decoding step $l-1$ and $h$ the encoder states for input $X$. In the following, we report the standard content-based mechanism and its location-aware variant which takes into account the alignment produced at the previous step using convolutional features:
		\begin{align*}
            e_{l, s} =  \begin{cases}
            				\text{content-based:}\\
                            \qquad w^T\, \tanh(W d_{l - 1} + Vh_{s} + b)\\
                        	\text{location-based:}\\
                            \qquad f_{u} = F \star \alpha_{ - 1}\\
                            \qquad w^T\, \tanh(W d_{l - 1} + Vh_{s} + Uf_{l, s} + b)
                        \end{cases}
        \end{align*}
where $w$ and $b$ are vectors, $W$ the matrix for the decoder, $V$ the matrix for the high representation $h$ and $U$ the matrix for the convolutional filters, that takes the previous alignment for location-based attention mechanism into account.

    \subsection{RNN transducer}\label{transducer}
    
The RNN transducer architecture was first introduced by Graves and al. \cite{Graves12-STW} to address the main limitation of the proposed CTC network: it cannot model interdependencies as it assumes conditional independence between predictions at different time steps.\\
To tackle this issue, the authors introduced a CTC-like network augmented with a separate RNN network predicting each label given the previous ones, analogous to a language model. With the addition of another network taking into account both encoder and decoder outputs, the system can jointly model interdependencies between both inputs and outputs and within the output label sequence.

Although the CTC and RNN-transducer are similar, it should be noted that unlike CTC which represent a loss function, RNN-transducer defines a model structure composed of the following subnetworks :
\begin{itemize}
    \item The encoder or transcription network: from an input value $x_{t}$ at timestep $t$ this network yields an output vector $h_{t}$ of dimension $|\mathcal{A}+1|$, where $+1$ denotes the ${<}blank{>}$ label which acts similarly as in CTC model.
    \item The prediction network: given as input the previous label prediction $y_{u-1} \in \mathcal{A}$, this network compute an output vector $d_{u}$ dependent of the entire label sequence $y_{0}, ..., y_{u-1}$.
    \item The joint network: using both encoder outputs $h_{t}^{enc}$ and prediction outputs $d_{u}^{dec}$, it computes $z_{t,u}$ for each input $t$ in the encoder sequence and label $u$ in prediction network such as:
        \begin{align*}
            h_{t, u}^{joint} &= tanh(h_{t}^{enc} + h_{u}^{dec})\\
            z_{t,u} &= lin(h_{t,u}^{joint})
        \end{align*} 
\end{itemize}

The output from the joint network is then passed to a softmax layer which defines a probability distribution over the set of possible target labels, including the blank symbol.

It should be noted that we made a small modification compared to the last proposed version \cite{Graves13-SPW}: instead of feeding the hidden activations of both networks into a separate linear layer, whose outputs are then normalised, we include another linear layer and feed each hidden activations to its corresponding linear layer which yields a vector of dimension $J$, the defined \textit{joint-space}.

Similarly to the CTC, the marginalized alignments are local and monotonic and the label likelihood can be computed using dynamic programming. However, unlike CTC, RNN transducer allows prediction of multiple characters at one time step, alongside their vertical probability transitions.
    \subsection{Other notable approaches}

    	\textbf{Joint CTC-attention\,}\label{joint_ctc_att}
The key idea behind the joint CTC-Attention \cite{Kim17-JCA} learning approach is simple. By training simultaneously the encoder using the attention mechanism with a standard CTC objective function as an auxiliary task, monotonic alignments between speech and label sequences can be enforced to reduce the irregular alignments caused by large jumps or loops on the same frame in the attention-based model. 
The objective function below formulates the multi-task learning of the network, where $0\leq\lambda\leq1$ is a tunable parameter weighting the contribution of each loss function:
            \begin{align*}
            	\mathcal{L}_{MTL} &= \lambda \mathcal{L}_{\textrm{ctc}} + (1 - \lambda) \mathcal{L}_{\textrm{att}}\\
                				  &= \lambda\, \text{log}\, p_{\textrm{ctc}}(Y|x) + (1 - \lambda)\, \text{log}\, p_{\textrm{att}}(Y|x)
            \end{align*}
The approach proposed in \cite{Hori17-AIJ} introduced a joint-decoding method  to take into account the CTC predictions in the beam-search based decoding process of the attention-based model. Considering the difficulty to combine their respective scores, the attention-based decoder performs the beam search character-synchronously whereas the CTC performs it frame-synchronously, two methods were proposed. 

The first one is a two-pass decoding process where the complete hypotheses from the attention model are computed and then rescored according to the following equation, where $p_{\textrm{ctc}}(Y|x)$ is computed using the standard CTC forward-backward algorithm:  
    \begin{align*}
    	\hat{Y} = \text{arg} \max\limits_{C \in A*} \{\lambda\, \text{log }\, p_{\textrm{ctc}}(Y|x) + (1 - \lambda)\, \text{log }\, p_{\textrm{att}}(Y|x)\}
    \end{align*}
The second method is a one-pass decoding method where the probability of each partial hypothesis in the beam search process is computed directly using both CTC and attention model such as, given $h$ the partial hypothesis and $\alpha$ the score defined as the log probability of the hypothesized sequence:  
    \begin{equation*}
    	\alpha (h) = \lambda \alpha_{\textrm{ctc}}(h) + (1 - \lambda) \alpha_{\textrm{att}}(h)
    \end{equation*}
        \textbf{End-to-end lattice-free MMI\,}
The end-to-end Lattice-Free MMI \cite{Hadian18-ETE} is the end-to-end version of the method introduced by Povey et al. \cite{Povey16-PST}.
In this version, a flat-start manner is adopted in order to remove the need of training an initial HMM-GMM for alignments and the tree-building pipeline.
Although the approach seems more like a flat-start adaptation of the state-of-art method than end-to-end in terms of pipeline and it does not benefit from the open-vocabulary property to construct unseen words compared to previously presented methods, we use it in our experiments as it showed small degradation over the original lattice-free MMI with different acoustic units.
We can therefore contrast the orthographic differences in productions between open systems and more constrained ones where the relationship between acoustic units and a word-level representation is restricted.

    \textbf{RNN-transducer with attention\,}
The RNN transducer architecture augmented with attention mechanisms was first mentioned, to the best of our knowledge, in \cite{Pra17-CSS}. Here, the prediction network described in \ref{transducer} is replaced by an attention-based decoder similar to the one described in \ref{attention} and used in the joint CTC-attention. This modification allows the decoder to access acoustic information alongside the sequence of previous predictions. As the decoder output computation is not affected by this change (the decoder and joint outputs computation are not dependent on a particular choice of segmentation), the architecture can be trained with the same forward-backward algorithm used for standard RNN-transducer. Finally, unlike the previous hybrid procedure, the inference procedure can be performed frame-synchronously with an unmodified greedy or beam search algorithm.

\section{Database}

We carried out our experiments using the data provided during the ESTER evaluation campaign (\textit{Evaluation of Broadcast News enriched transcription systems}) \cite{Galliano09-TE2} which is one of the most commonly used corpus for the evaluation of French ASR.
Evaluations are done on test set. The details of the dataset, corresponding to 6h34 of speech, are described in \cite{Galliano09-TE2}. We use the same normalization and scoring rules as in the evaluation plan of the ESTER 2 campaign except that we do not use equivalence dictionary and partially pronounced words are scored as full words.

To train the acoustic models we use the 90h of the training set from ESTER2 augmented by 75h from ESTER1 training set and 90h from the additional subset provided in ESTER1 with their transcriptions provided in the corpus EPAC \cite{Esteve10-TEC}. We removed segments containing less than 1,5 seconds of transcribed speech and we excluded the utterances corresponding to segments with more than 3000 input frames or sentences of more than 400 characters for the end-to-end models.
%Some irregulars segment-utterance pairs remaining, we re-segmented the training data using a GMM-HMM model with LDA-MLLT-SAT features we build our phone-based chain model upon. 
Because some irregulars segment-utterance pairs remained, we re-segmented the training data using the GMM-HMM model (with LDA-MLLT-SAT features) we build our phone-based chain model upon. 
During re-segmentation, only the audio parts matching the transcripts are selected. This brings the training data to approximately 231h. For neural networks training, we have applied 3-fold speed perturbation \cite{Ko15-AAF} and volume perturbation with random volume scale factor between 0.25 and 2, leading to a total of training data of 700h.

For language modeling, we use the manual transcripts from the training set. We extend this set with manually selected transcriptions from other speech sources (BREF corpus \cite{Larnel91-BAL}, oral interventions in EuroParl from '96-'06 \cite{Koehn05-APC} and a small portion of transcriptions from internal projects). The final corpus is composed of 2 041 916 sentences, for a total of 46 840 583 words.

\section{Implementations}

All our systems share equivalent optimization -- no rescoring technique or post-processing is done -- as well as equivalent resource usage. Each system is kept to its initial form (\textit{i.e.} no further training on top of the reported system).

    \subsection{Acoustic units}\label{units}
    
For our experiments, three kind of acoustic units were chosen: phones, characters and subwords. The baseline phone-based systems use the standard 36 phones used in French.
The CTC, attention and hybrid systems each have two versions: one for characters with 41 classes (26 letters from the Latin alphabet, 14 letters with a diacritic and apostrophe) and another version for subwords where the number of classes is set to 500, the final set of subword units used in our training being selected by using a subword segmentation algorithm based on a unigram language model \cite{Kudo18-SRI} and implemented in Google's toolkit SentencePiece \cite{Kudo18-SPA}. For the end-to-end variant of the chain model, characters units are used with the 41 classes set.

    \subsection{Baseline systems}\label{baseline}
    
We used the Kaldi toolkit \cite{Povey11-TKS} to train the chain model and its end-to-end variant.

The chain model is a TDNN-HMM model trained with the LF-MMI objective function. The neural network is based on a sub-sampled time-delay neural network (TDNN) with 7 TDNN layers and 1024 units in each, time stride value being set to 1 in the first three layers, 0 in the fourth layer and 3 in the final ones.
The end-to-end version of the chain model is trained in the same way as the original model but with a different architecture. The network is composed of a 1 LSTM Projected layer \cite{Sak14-LST} with 512 units followed by 2 TDNN layers of 512 units - these first three layers being repeated twice - and another 1 LSTM Projected layer with 512 units when using character as unit. The time delay value in the recurrent connections of the projected LSTM layers is set to 3.

As the input for our models, we use a 40-dimensional high resolution MFCC vector (\textit{i.e.} linear transform of the filterbanks) and CMVN for both the chain model trained with lattice free-MMI and its end-to-end variant. We also trained separately a phone-based chain model with the previous 40-dimensional MFCC vector concatenated with a 100-dimensional i-vector \cite{Gupta14-IVB} as input to assess the impact of speaker-dependant features.

For the linguistic part, we also trained a word 3-gram language model using SRILM's n-gram counting method \cite{Stolcke02-SRI} with KN discounting. As lexicon we use the phonetic dictionary provided by the LIUM, thus the vocabulary of our language model is limited to the most frequent 50k words found in our training texts and also present in their dictionary.

For the end-to-end version modeling characters, we replace the phonetic lexicon by an orthographic lexicon with the same entries, where the orthographic representation is the word sequence with space inserted between each character. 

    \subsection{End-to-end systems}

We use the ESPNET toolkit \cite{Watanabe18-ESP} to train the five end-to-end systems. For each method two acoustic units are used: character and subword. Ten epochs are used to train each model. 

The acoustic models for all methods share the same architecture composed of VGG bottleneck \cite{Hori17-AIJ} followed by a 3-layer bidirectional LSTM with 1024 units in each layer and each direction. For the models using attention mechanism we use a 1-layer LSTM with 1024 units and location-based mechanism with 10 centered convolution filters of width 100 for the convolutional feature extraction as decoder. When training jointly CTC and attention, $\lambda$ was set to $0.3$ based on preliminary experiments. For RNN-transducer the joint space between encoder and decoder was set to 1024 dimensions.\\
The input features for these models are a 80-dimensional raw filterbanks vector with their first and second derivatives with cepstral mean normalization (CMN).

For the experiments involving language models, we trained three different models using the RNNLM module available in ESPNET: one with characters, another with subwords and the last one with full words for multi-level combination when dealing with characters as units. Each model is incorporated at inference time using shallow fusion \cite{Kannan18-AAI}, except for the word-LM relying on multi-level decoding \cite{Hori17-MLL}. The main architecture of our RNNLMs is a 1-layer RNN, the number of units in each layer depending of the target unit: 650 units for subwords and characters, and 1024 units for words. Unlike the systems described above, the vocabulary for the word-based RNNLM was limited according to the training texts only.

In order to directly compare the baseline systems to the end-to-end systems relying on different word-based LM (\textit{i.e.} N-gram and RNN-based), another RNNLM was trained using available tools in Kaldi. The language model shares the same architecture as the word-RNNLM described in this subsection and was trained with equivalent training parameters. Following lattice rescoring approach proposed in \cite{Xu18-APR}, decoding was then performed with the RNNLM for all baseline systems. We observe a maximal WER improvement of 0.12\% on the dev set and 0.16\% on the test set compared to the systems relying on the original 3-gram. Adding to that a difference of less than 1.3\% between words in language model vocabularies for baseline and end-to-end systems, we thus consider minimal the impact for our comparison.

    \subsection{Decoding}

To measure the best performance, we set the beam size to 30 in decoding under all conditions and for all models.
When decoding with the attention-only model, we do not use sequence length control parameters such as coverage term or length normalization parameters \cite{Wu16-GNM}.
When joint-decoding, $\lambda$ is set to 0.2 based on our preliminary experiments.
For CTC and attention experiments involving a RNNLM, the language model weight during decoding is set to respectively $0.3$ for character and subword LM, and $1.0$ for the word LM. For RNN-transducer, we downscale the use of external language model when performing multi-level LM decoding, setting the value to $0.3$.

    \section{Results}\label{results}
            \begin{table*}[t]
        	\caption{Character Error Rate (CER) with detailed report and Word Error Rate (WER) for all evaluated methods on ESTER testing set. Italic values denotes errors on subwords. Bold values indicates best results for each section.}
        	\label{table:cer_wer_results}
            \footnotesize
           	\centering
        	\begin{tabular}{c|c|c|c|c|c|c|c|c||c|}
  	 	 	\bfseries{Model} & \bfseries{Units} & \bfseries{Lexicon} & \bfseries{LM} & \bfseries{Corr.} & \bfseries{Sub.} & \bfseries{Del.} & \bfseries{Ins.} & \bfseries{CER} & \bfseries{WER}\\\thickhline\thickhline
 			chain LF-MMI & \multirow{3}*{phone} & \multirow{3}*{50K} & \multirow{3}*{word 3-gram} & \cellcolor{gray!25} & \cellcolor{gray!25} & \cellcolor{gray!25} & \cellcolor{gray!25} & \cellcolor{gray!25} & 14.2\\\cline{1-1}\cline{10-10}
            \begin{tabular}{@{}c@{}}chain LF-MMI\\\textit{(i-vectors)}\end{tabular} & & & & \cellcolor{gray!25} & \cellcolor{gray!25} & \cellcolor{gray!25} & \cellcolor{gray!25} & \cellcolor{gray!25} & \textbf{13.7}\\\thickhline\thickhline
            \multirow{3}*{e2e chain LF-MMI} & phone & \multirow{3}*{50K} & \begin{tabular}{@{}c@{}}phone 4-gram\\+ word 3-gram\end{tabular} & \cellcolor{gray!25} & \cellcolor{gray!25} & \cellcolor{gray!25} & \cellcolor{gray!25} & \cellcolor{gray!25} & 14.4\\\cline{2-2}\cline{4-9}\cline{10-10}
            & char & & \begin{tabular}{@{}c@{}}char 4-gram\\+ word 3-gram\end{tabular} & \textbf{94.3} & 2.6 & \textbf{2.0} & 3.0 & \textbf{7.6} & 14.8\\\cline{10-10}\thickhline\thickhline
            \multirow{6}*{CTC} & \multirow{3}*{char} & \multirow{2}*{None} & None & 87.4 & 4.9 & 7.7 & 3.0 & 15.5 & 42.3\\\cline{4-10}
             & & & char RNNLM & 89.5 & 4.4 & 6.1 & 2.8 & 13.3 & 31.0\\\cline{3-10}
             & & 50K & word RNNLM & 89.8 & 4.3 & 5.8 & 2.7 & 12.8 & 27.3\\\cline{2-10}
             & \multirow{2}*{subword} & \multirow{2}*{None} & None & \textit{81.2} & \textit{9.2} & \textit{9.6} & \textbf{\textit{1.4}} & \textit{20.1} & 28.4\\\cline{4-10}
             & & & subword RNNLM & \textit{85.7} & \textit{9.1} & \textit{6.1} & \textit{2.3} & \textit{17.5} & 21.2\\\thickhline
             \multirow{5}*{\begin{tabular}{@{}c@{}}Attention\\\textit{(location-based)}\end{tabular}} & \multirow{3}*{char} & \multirow{2}*{None} & None & 89.8 & 3.2 & 6.7 & 3.3 & 13.2 & 24.4\\\cline{4-10}
             & & & char RNNLM & 90.1 & 3.2 & 6.4 & 3.2 & 12.8 & 23.6\\\cline{3-10}
             & & 50K & word RNNLM & 90.2 & 3.3 & 6.2 & 3.2 & 12.7 & 23.0\\\cline{2-10}
             & \multirow{2}*{subword} & \multirow{2}*{None} & None & \textit{84.1} & \textit{12.3} & \textit{3.6} & \textit{3.6} & \textit{19.5} & 22.7\\\cline{4-10}
             & & & subword RNNLM & \textit{85.0} & \textit{11.1} & \textit{3.4} & \textit{3.3} & \textit{18.4} & 21.8\\\thickhline
            \multirow{6}*{RNN-transducer} & \multirow{3}*{char} & \multirow{2}*{None} & None & 93.9 & 2.8 & 3.3 & 2.4 & 8.5 & 19.7\\\cline{4-10}
             & & & char RNNLM & 94.0 & 2.6 & 3.4 & 2.2 & 8.2 & 18.8\\\cline{3-10}
             & & 50K & word RNNLM & 94.1 & 2.5 & 3.4 & \textbf{2.1} & 8.0 & 18.1\\\cline{2-10}
             & \multirow{2}*{subword} & \multirow{2}*{None} & None & \textit{87.1} & \textit{8.9} & \textit{4.1} & \textit{2.5} & \textit{15.5} & 18.5\\\cline{4-10}
             & & & subword RNNLM & \textbf{\textit{87.4}} & \textbf{\textit{8.2}} & \textit{4.3} & \textit{2.2} & \textit{14.7} & 17.4\\\thickhline\thickhline
             \multirow{5}*{\begin{tabular}{@{}c@{}}Joint CTC-Attention\\\textit{+ joint decoding}\\\textit{\scriptsize(MTL=0.3, ctc weight=0.2)}\end{tabular}} & \multirow{3}*{char} & \multirow{2}*{None} & None & 91.7 & 2.9 & 5.4 & \textbf{2.1} & 10.4 & 22.1 \\\cline{4-10}
             & & & char RNNLM & 92.2 & 2.9 & 5.0 & 2.2 & 10.1 & 20.6\\\cline{3-10}
             & & 50K & word RNNLM & 92.8 & 3.1 & 4.1 & 2.4 & 9.6 & 18.6\\\cline{2-10}
             & \multirow{2}*{subword} & \multirow{2}*{None} & None & \textit{87.3} & \textit{9.3} & \textit{3.5} & \textit{2.5} & \textit{15.3} & 18.7 \\\cline{4-10}
             & & & subword RNNLM & \textbf{\textit{87.4}} & \textit{8.8} & \textbf{\textit{3.3}} & \textit{2.4} & \textbf{\textit{14.5}} & 17.8\\\thickhline
             \multirow{5}*{\begin{tabular}{@{}c@{}}RNN-transducer w/ attention\\\textit{(location-based)}\end{tabular}} & \multirow{3}*{char} & \multirow{2}*{None} & None & 94.1 & 2.7 & 3.2 & 2.3 & 8.2 & 19.1\\\cline{4-10}
             & & & char RNNLM & 94.1 & 2.5 & 3.4 & \textbf{2.1} & 8.0 & 18.3\\\cline{3-10}
             & & 50K & word RNNLM & \textbf{94.3} & \textbf{2.4} & 3.3 & \textbf{2.1} & 7.8 & 17.6\\\cline{2-10}
             & \multirow{2}*{subword} & \multirow{2}*{None} & None & \textit{87.1} & \textit{9.0} & \textit{4.1} & \textit{2.5} & \textit{15.6} & 18.4\\\cline{4-10}
             & & & subword RNNLM & \textit{87.3} & \textit{8.3} & \textit{4.4} & \textit{2.2} & \textit{14.9} & 17.5\\\thickhline
            \end{tabular}
    	\end{table*}

The results of our experiments in terms of Character Error Rate (CER) and Word Error Rate (WER) on the test set are gathered in the Table \ref{table:cer_wer_results}. For CER we also report errors in the metric: correct, substituted, inserted and deleted characters.\\
It should be noted that the default CER computation in all frameworks does not use a special character for space during scoring. As important information relative to this character, denoting word-boundary errors, can be observed through the WER variation during comparison, we kept the initial computation for CER. Thus, for low CER variations, bigger WER differences are expected notably between traditional and end-to-end systems.

\subsection{Baseline systems}

The phone-based chain model trained with lattice-free MMI criterion has a WER of 14.2 on the test set. Compared to the best reported system during the ESTER campaign (WER 12.1\% \cite{Galliano09-TE2}), the performance show a relative degradation of 14.8\%. Although the compared system rely on a HMM-GMM architecture, it should be noted that a triple-pass rescoring (+ post-processing) is applied, a consequent number of parameters is used, and a substantial amount of data is used for training the language model (more than 11 times our volume). Adding i-vectors features the performance of our model is further improved, leading to a WER of 13.7.

For the \textit{end-to-end} phone-based system we denote a small WER degradation of 0.2\% compared to the original system without i-vectors, which is a good trade-off considering the removal of the initial HMM-GMM training. Switching to characters as acoustic units we obtain a WER of 14.8, corresponding to a CER of 7.6. The detailed report show that all types of errors are quite balanced, with however a higher number of deletions. The system remains competitive even with orthographic units, despite the low correspondence between phonemes and letters in French. On the same note, a plain conversion of phonetic lexicon to a grapheme-based one does not negatively impact the performances. This was not excepted considering the use of alternative phonetic representation in French to denotes possible \textit{liaisons} (the pronunciation of the final consonant of a word immediately before a following vowel sound in preceding word).

\subsection{End-to-end systems}\label{results-ete}

\textbf{Character-based models\,} While, without language model, the attention-based model outperforms CTC model as expected, RNN-transducer performances exceed our initial estimations, surpassing previous models in terms of CER and WER. RNN-transducer even outperforms these models coupled with language model, regardless of the level of knowledge included (character and word-level). The CER obtained with this model is 8.5 while the WER is 19.7. This represent a relative decrease of almost 40\% for the CER and 17\% for the WER against the attention-based model with word LM, the second best system for \textit{classic} end-to-end. Compared to the end-to-end chain model system modeling characters, we observe a small CER difference of 0.9 which corresponds to a WER difference of 4.9. While the CER is competitive, errors at word-level seem to indicate difficulties to model word boundaries compared to baseline systems.

Extending the comparison to hybrid models, only the RNN-transducer with attention mechanism could achieve similar or better results than its vanilla version. Although the joint CTC-attention procedure is beneficial to correct some limitations from individual approaches, the system can only reach a CER of 10.4 equivalent to a WER of 22.1. However, by adding word LM and using multi-level decoding, the system can achieve closer WER performance (18.6) despite the significant difference in terms of CER (9.6).\\
For the hybrid transducer relying on additional attention module, performances in all experiments are further improved compared to standard, reaching 8.2\% CER and 19.1\% WER without language model.

Concerning the best systems, it should be noted that the RNN-transducer performance is further improved with the use of language model, obtaining a CER of 8.0, close to our baseline score (7.6), with a word LM. In terms of WER it represents a relative improvement of 8.5\% against previous results, which is however still far from the performance denoted with the baseline system for this metric (14.8\%). For the RNN-transducer with an attention decoder, we achieve even better performance with a CER of 7.8 equivalent to a WER of 17.6. This is our best model with characters as \textit{acoustic} units.

Focusing on the CER report, several observations can be made :

Insertion errors are lower for CTC models than attention-based systems, with the addition of language models included. Attention-based are expected to have higher number of deletions or insertions depending of the length difference between input and output sequences, it is however unanticipated to observe such a high number of deletion errors.

Following the last observation, we investigated the deletion errors done by the attention-only model. From what we found, the main reason is the existence of irregular segment-utterance pairs in the dataset (i.e: really low correspondence). Using coverage, penalty or length ratio terms helped on problematic pairs but degraded the global performances, regular short or long pairs being impacted.

Adding a language model decreases all errors in CTC systems while only deletion errors decrease in the attention-only system. Coupled with a word language model, substitution errors are even higher for attention model.

Similar observations can be made for RNN transducer. While we observe a small decrease of insertion errors with the addition of a language model, we also see a small increase in deletion errors. However the system is more impacted by the insertion changes as the number of substitutions decrease and the number of correct words increase.

Despite similar CER performances between CTC model with word LM and attention-only model with character or word LM for example, the first system cannot reach the word error rate of the second systems. It is beneficial to model linguistic information alongside acoustic information rather than in an external language model being at character or word level, although both can be combined to reach better performances. However, we should also consider that the training data for the acoustic model is the same as the data used to train the LM, augmented with a volume equivalent to less than a quarter of the initial training sentences.

Comparing the end-to-end chain model modeling characters to the RNN-transducer with language models, we can extract several useful information. Deletion errors made by the transducer are more influential at word level than the insertion errors made by the baseline system. From the hypotheses we observed that the insertion errors mostly happen on ambiguous verbal forms, gender forms or singular/plural forms in the baseline system. For the transducer, the same behaviour is observed however deletion errors at character level mostly happen on small words (such as article), common names and proper names which are numerous in the corpus.

Although we observe a smaller number of substitutions at character level for the RNN-transducer with or without attention compared to the baseline system, substitution errors impact more words than the baseline system. These errors are mostly due to the same problems described previously, while substitutions in baseline systems are more localized due particularly to the presence of OOV and ambiguous words.

Considering all the previous observations, further investigation should be done to compare and categorize errors at character and word level in each system and also assess the value of these errors. The character errors reported for RNN-transducer with attention should be sufficient motivation as we report, against the baseline system, a lower number of substitution and insertion errors coupled to an equivalent number of correct words despite a significant gap in WER performance.

\textbf{Subword-based models\,} Replacing characters with subword units improves the overall performance of all end-to-end methods. The gain is particularly important for CTC lowering the WER from $42.3$ to $28.4$ without language model. The gain observed when adding the language model to CTC is impressive with a relative improvement of almost 28\% on WER (from $28.4$ to $21.2$). For the system relying only on attention, the WER is further improved without and with language model but, unlike when we used characters, the model is outperformed on both CER and WER by the model relying on CTC. Although we observe a similar CER for both methods we also note a significant difference in terms of correct characters and WER (almost $6\%$). The attention making mostly consecutive mistakes on the same words or groups of words (particularly at the beginning and end of utterances) while the CTC tends to recognize part of words as independent, thus incorrectly recognizing word boundaries. Adding RNN-transducer to the comparison, both previous methods are surpassed, on CER ($20.1$ for CTC, $17.5$ for attention and $15.2$ for transducer) and on WER ($21.1$ for CTC, $21.8$ for attention and $18.4$ for transducer). Decoding with an external language model, the CER and WER are further improved by about $5.5\%$ and $6.0\%$. It should be noted that the transducer model without language model exceed CTC and attention coupled to the subword LM.

Adding the hybrid systems to the comparison, we denote some differences compared to character-based systems. The RNN-transducer is not improved with attention mechanism and even slightly degraded for both CER and WER. The same observations can be done with and without LM addition. It seems the attention mechanism has more difficulty to model intra-subwords relations than intra-characters relations. However further work should be allocated to extend the  comparison with different attention mechanisms, such as multi-head attention, and estimate the influence of architecture depending on output dimensions and representations.\\
Concerning the last hybrid system, joint CTC-attention is better suited to subword than characters, reaching comparable performances to transducer even without language model: 18.7\% against $18.4$ for RNN-transducer and $18.5$. Although transducer are reported as our best system, it should be noted that joint CTC-attention reach equal or better performance on subword errors. Talking only about conventional ASR metric, we consider the two hybrid systems and vanilla transducer equivalent for subword units.

As in the previous section, we also made a focus on the detailed error report and denoted some differences compared to previous observations:

Akin to previous observations with characters, insertion errors are lower for CTC models ($1.4\%$) than attention-based models ($3.6\%$) with subwords. However, here, the number of insertions for CTC is even lower than for all other methods, transducer and hybrid systems showing an average insertion error of $2.5\%$.

Previously, we noted that a higher number of deletions or insertions should be expected with attention-only model. With subword units, we can observe a balanced number of deletions and insertions although we also denote a significant number of substitutions. Following this new observation, we also investigated the orthographic output from both models. We denoted that the limitation of attention model was mostly removed and word sequence was unrolled or stopped. However it translated to a really large number of substitutions, some subwords within the word structure being repeated or cut.

Although, we report a higher number of correct words and a lower number of errors for joint CTC-attention, the hybrid method obtains a higher WER than RNN-transducer and its hybrid version. Analyzing the hypothesis formulated and error distribution by both systems we could not extract any relevant information to explain the number of words impacted by the errors at character level.

On the same note, the following difference should still be noted: transducer-based models have a lower number of substitutions and equivalent or lower insertion whereas joint CTC-attention has a lower number of deletions and an equivalent or higher number of correct characters. Outside correct labels, only the CTC has a similar error distribution.

In case of joint CTC-Attention we can see that CTC as auxiliary function brings some benefices: the number of substitutions and insertions being further reduced compared to attention-only model. Additionally, the number of deletions is kept to the same range despite a high number of deletions for the CTC-only model. In case of additional attention module for RNN-transducer, although the attention-only has a lower number of deletion errors ($3.6$ versus $4.1$ for RNN-transducer), the inclusion of attention mechanism did not help to reduce this number. The error distribution is the same with and without attention. It also should be noted that RNN-transducer with attention has equivalent performance with characters and subword units.

Adding language models, all errors are lowered. The only exceptions being the number of insertions for CTC ($1.4\%$ raised to $2.3$), the number of deletions for RNN-transducer (from $4.1\%$ to $4.3$) and its hybrid counterpart (from $4.1$ to $4.4$). In these cases, and similarly as when we use character units, we can observe that the error rate (e.g.: insertion) decreases when the other (e.g.: deletion) increases. 

\section{Conclusion}

In this paper, we experimentally showed that end-to-end approaches and different orthographic units were rather suitable to model the French language. RNN-transducer was found specially competitive with character units compared to other end-to-end approaches. Among the two orthographic units, subword was found beneficial for most methods to address the problems described in section \ref{results-ete} and retain information on ambiguous patterns in French. Extending with language models, we could obtain promising results compared to traditional phone-based systems. The best performing systems being for character unit the RNN-transducer with additional attention module, achieving 7.8\% in terms of CER and 17.6\% on WER. For subword units, classic RNN-transducer, RNN-transducer with attention and joint CTC-attention show comparable performance on subword error rate and WER, with the first one being slightly better on WER ($17.4\%$) and the last one having a lower error rate on subword ($14.5\%$).\\
However, we also showed difference in produced errors for each method and different impact at word-level depending of the approach or units. Thus, future work will focus on analysing the orthographic output of these systems in two ways: 1) investigate errors produced by the end-to-end methods and explore several approaches to correct common errors done in French and 2) compare the end-to-end methods in a SLU context and evaluate the semantic value of the \textit{partially} correct produced words.

%\newpage
%\vspace{-0.1cm}
\bibliographystyle{IEEEtran}
\normalsize
\bibliography{mybib}
\vfill

\end{document}